\documentclass[10pt,twocolumn,letterpaper]{article}
\usepackage[accsupp]{axessibility} 
\usepackage{iccv}
\usepackage{times}
\usepackage{epsfig}
\usepackage{graphicx}
\usepackage{amsmath}
\usepackage{amssymb}
\usepackage{microtype}
\usepackage{graphicx}
\usepackage{subcaption}
\usepackage{amsmath,amsfonts,amssymb}
\usepackage{algorithm}
\usepackage{algorithmic}
\usepackage{nicefrac}
\usepackage{multirow}
\usepackage{diagbox}
\usepackage{booktabs} % for professional tables
\usepackage{stfloats}
\usepackage[pagebackref=true,breaklinks=true,letterpaper=true,colorlinks,bookmarks=false]{hyperref}

\iccvfinalcopy % *** Uncomment this line for the final submission

\def\iccvPaperID{33} % *** Enter the ICCV Paper ID here

\ificcvfinal\fi

\begin{document}

%%%%%%%%% TITLE
\title{DeepFake MNIST+: A DeepFake Facial Animation Dataset}

\author{\small Jiajun Huang\\
\small The University of Sydney\\

\and
\small Xueyu Wang\\
\small The University of Sydney\\
\and
\small Bo Du\\
\small Wuhan University\\
\and
\small Pei Du\\
\small AntGroup\\
\and
\small Chang Xu\\
\small The University of Sydney\\

}

\maketitle
\vspace*{-1.5cm}
\ificcvfinal\thispagestyle{empty}\fi

%%%%%%%%% ABSTRACT
\begin{abstract}
The DeepFakes, which are the facial manipulation techniques, is the emerging threat to digital society. Various DeepFake detection methods and datasets are proposed for detecting such data, especially for face-swapping. However, recent researches less consider facial animation, which is also important in the DeepFake attack side.  It tries to animate a face image with actions provided by a driving video, which also leads to a concern about the security of recent payment systems that reply on liveness detection to authenticate real users via recognising a sequence of user facial actions. However, our experiments show that the existed datasets are not sufficient to develop reliable detection methods. While the current liveness detector cannot defend such videos as the attack. As a response, we propose a new human face animation dataset, called DeepFake MNIST+\footnote{https://github.com/huangjiadidi/DeepFakeMnist}, generated by a SOTA image animation generator. It includes 10,000 facial animation videos in ten different actions, which can spoof the recent liveness detectors. A baseline detection method and a comprehensive analysis of the method is also included in this paper. In addition, we analyze the proposed dataset's properties and reveal the difficulty and importance of detecting animation datasets under different types of motion and compression quality.
\end{abstract}

%%%%%%%%% BODY TEXT
\section{Introduction}
DeepFake\footnote{DeepFake not only indicates the facial modification methods but also is the name of one algorithm called deepfake~\cite{GitHubdf58:online}.} has become a critical topic for our digital society. With DeepFake, we can now easily change the identity or expression of the face in an image or a piece of video with another person's identity or expression. The mainstream DeepFake techniques are based on Deep Neural Networks (DNNs), especially Generative Adversarial Networks (GANs)~\cite{goodfellow2014generative}, to produce visually plausible images or videos which are hard to be discriminated by humans. There is growing concern about DeepFake, as malicious people could use the techniques to palm off the victims and illude presence and activities, even if they never did before.

A number of DeepFake methods have been developed to manipulate the attributes of human face in images or videos. For example, the swapping methods~\cite{GitHubip51:online, GitHubdf58:online, GitHubde97:online, GitHubsh44:online, FakeApp295:online, li2020advancing} mostly focus on the identity of the face and try to replace the face in one image/video with the face from others. The deepfake method~\cite{GitHubdf58:online}, a famous swapping algorithm, trains identity-dependent two auto-encoders to swap the faces of two identities. Besides face identity, many other face attributes have been studied in the literature. For example, \cite{thies2016face2face, thies2019deferred} modified the expression in one face image/video, and facial image animation~\cite{bansal2018recycle, wang2018video, siarohin2020first, suwajanakorn2017synthesizing, burkov2020neural}, as a compose of expression manipulation, is increasingly being important within the DeepFakes. Given a face source image and the driving video, DeepFake can now generate a new video where the source face performs the same action as the driving video. For instance, Siarohin et al.~\cite{siarohin2020first} use an encoder to capture optical flow information from the videos, embedding the information with source images, and generate videos. Zakharov et al.~\cite{zakharov2019few} pass the identity embedding to the image generator to produce manipulated face with the given landmark. Burkov~\cite{burkov2020neural} extracts the identity and pose information separately and generate videos with embedding.

Given the growing anxiety on the high-quality generation by DeepFake and its potential negative social impacts, it becomes especially urgent to study the defense techniques against DeepFake. Recently a few datasets have been created for the study of DeepFake detection methods. UADFV~\cite{yang2019exposing} and Celeb-DF~\cite{li2020celeb} collect youtube videos to generate face-swapping videos, and DFDC~\cite{dolhansky2020deepfake} captures 48,190 videos with paid actors and generates a large scale swapping dataset with over 104,500 videos. By analyzing these datasets,  Rossler et al.~\cite{rossler2019faceforensics++} and Seferbekov~\cite{GitHubse90:online} suggested the importance of CNN architectures on detecting the swapped faces. Li et al.~\cite{li2020face} further found the manipulate boundary between the face and head can be an effective clue for the detection. However, all these works are mostly about detecting the face identity change, which is a limitation of existing deepfake datasets. There is rare dataset or deepfake detection work about facial animation, though it occupies a significant part in DeepFake attack side.

% However, it has fewer works about facial animation. The only related public dataset is provided from FF++ DF~\cite{rossler2019faceforensics++}, which is generated by two expression manipulation methods. 

Recently, facial animation by DeepFake has been deployed on mobile devices, producing a large number of fake videos and broadcasts through the Internet~\cite{Avatarif82:online}. These fake videos thus challenge the security of many intelligent systems in our daily life. For example, the face recognition based payment systems usually rely on liveness detection to verify whether users are the real people by requiring them to do a sequence of specific actions in videos. However, our experiments show that DeepFake detectors trained on existing deepfake datasets consisting of face identity changes are not applicable for detecting facial animation videos. Further,  we observe that the SOTA liveness detector in public cannot defend the animation data with specific actions as they claimed. 
%It is urgent for us to create an action-specific facial animation video dataset to detect and defend the animation videos.

In this paper, we propose a new dataset, called DeepFake MNIST+, a human face animation video dataset. The DeepFake MNIST+ dataset is developed as a response to the wide use of the MNIST dataset, but for a different deepfake detection problem. We create this dataset to provide the basis for learning and practicing how to develop, evaluate, and use deep neural networks for fake face animation detection. The dataset contains 10,000 face animation videos in ten different actions, \emph{plus} 10,000 real face videos to enable a supervised detector training. These fake facial animation are of higher fidelity and able to spoof the popular liveness detectors on the market (as of the time of this manuscript submission). Given the gap of different deepfake types, the detectors trained on existing DeepFake datasets with face identity change cannot well detect fake animations in the proposed dataset. We  establish deepfake detection baselines on the DeepFake MNIST+ dataset and carefully evaluate their performance in different scenarios.

We also present comprehensive analysis related to the properties of proposed dataset. We explore the impact of motion type and compression quality of generated videos. As observed from Figure \ref{fig::each_action_train_only}, the actions with large movements will challenge existing detectors. But by taking these difficult DeepFake data into account, the learned detector can enjoy a much better generalization across other types of DeepFake data. The relevant discussion and empirical studies in this paper therefore shed new light on both DeepFake generation and detection research.

\section{Related Works}
\subsection{Image Animation}
The interest in facial manipulation methods has rapidly increased recently. One approach for manipulation is based on 3D modeling. Zollhofer et al., ~\cite{zollhofer2018state} build a 3D morphable model for the source face, to perform realistic animation for given actions. Suwajanakorn ~\cite{suwajanakorn2017synthesizing} attempt to model lip to forgery talking. The deepfake~\cite{GitHubdf58:online} introduce a DNN based face-swapping method, replacing faces within two identities with two encoders. Although it requires plenty of videos/images of both identities to achieve better results, the promising results show the potential of manipulation with DNNs. The recent researches focus on identity-independent swapping methods. Li et al.~\cite{li2020advancing} implement two encoders to extract attribution and identity information and embed them in GAN to generate high fidelity swapped results. Several DNN based expression manipulation and animation are also proposed. Thies et al.~\cite{thies2019deferred} consider facial reenactment as a domain transfer problem using Pix2Pix architecture~\cite{isola2017image} to produce results. Siarohin et al.~\cite{siarohin2020first} extract the motion information of driving videos with optimal flow estimation to generate high-quality animation results.

\subsection{DeepFake and Liveness Detection}
An increasing number of DeepFake detection methods are proposed as a response to the huge concern from society. The approaches could be separated into three categories. The first type is trying to detect the unnatural section of the manipulated videos, such as swapping boundary~\cite{li2020face}, inconsistent head angles between the face and head~\cite{yang2019exposing}. The second type detects the synthesis signal of GAN to distinguish DeepFake data. For instance, Wang et al.~\cite{wang2020cnn} observe GAN signatures using discrete cosine transform for detecting CNN-based DeepFake samples. The last categories' approaches rely on DeepFake dataset to train the detectors~\cite{rossler2019faceforensics++, GitHubse90:online, guera2018deepfake, nguyen2019capsule}, regardless of the inconsistent or signals.

On the other hand, the recent liveness detector mainly defending the psychical level attacks called Replay Attacks. For instance, attackers could build a 3D mask of victims through a 3D printer or print out victims' face in the paper and wear it. To defense against such attacks, many detection methods have been proposed. Some approaches try to detect the differences between real faces and forgery faces. De et.al., ~\cite{de2012moving} estimate the invariant of facial points for detection. Komulainen ~\cite{komulainen2012face} believe detecting faces' dynamic muscle change can distinguish the spoofing. Wang et al., ~\cite{wang2017face} detect the blood flow change under the skin to separate facial mask and real face. The recent payment or identity verification solutions with smartpohone usually combine pose verification with liveness detection, such as Alipay. They usually require users to do some specific action, such as blinking or yawing, to improve the performance. However, our experiments show that the facial animation data with specific actions could spoof the SOTA liveness detectors in the market.

\subsection{DeepFake Datasets}

A few DeepFake datasets have been proposed recently as a response to the increasing concerns on DeepFake techniques as they can generate realistic results to spoof people. However, most of the datasets are generated using identity swapping algorithms, but only a few works for facial animation. Table~\ref{table:dataset_info} shows the details of these DeepFake datasets.

\textbf{Celeb-DF~\cite{li2020celeb}}: The Celeb-DF is a large face swap dataset. It selects 590 real videos from Youtube, which all about the talking videos of celebrities. The dataset contains 5,639 synthesis videos using a high-resolution face generator. 

\textbf{DFDC~\cite{dolhansky2020deepfake} }: The Facebook DeepFake detection challenge dataset is one of the largest face-swapping video dataset recently. It contains 104,500 face swap videos based on 48,190 source videos shot with 3,426 paid actors in different locations and light conditions.

\textbf{FF++~\cite{rossler2019faceforensics++}}: The FaceForensics++ dataset is one of the popular DeepFake datasets. The autoher use multiple DeepFake methods to generate the dataset. The dataset to provide the data generated by expression manipulation, the related techniques with facial animation. Two manipulation methods, Face2Face~\cite{thies2016face2face} and NeuralTextures~\cite{thies2019deferred}, are implemented to generated DeepFake videos, while the two face-swapping methods~\cite{GitHubip51:online, GitHubdf58:online} are also included. The dataset uses 1,000 real videos from Youtube to generate 1,000 DeepFake videos for each method.

the proposed dataset is a contemporaneous work with Forgerynet~\cite{he2021forgerynet}.  Our proposed dataset has several differences comparing with it: i) the proposed dataset includes animation data under ten specific categories (e.g., head movement and emotion changes), rather than applying the unknown action from a random video; ii) we boost the quality and challenge of the proposed dataset by filtering the generation with liveness detection; and iii) we present a comprehensive analysis about the proposed dataset, including action categories and video quality.

According to Table~\ref{table:dataset_info}, the only  dataset involving the face animation is FF++, which is a small dataset and is hard to cover the challenging deepfake data in the real world. We argue that the face animations in a deepfake dataset should be diverse enough and are better to cover the animation categories in the prospective downstream tasks, instead of the just causal talking in FF++. For example, the liveness detectors often require specific actions or expressions of the face as the input. For these reasons, it engages us to propose a large-scale and action-specific facial animation dataset.

\begin{table}[h]
\vspace{-10pt}
\centering
\small

\scalebox{0.85}{
% \def\arraystretch{0.8}%
%   \normalsize
\begin{tabular}{c|cccc}
\toprule
                                                                   & \#real & \#fake & \begin{tabular}[c]{@{}c@{}}type(s) of \\ generation\end{tabular}                      & \begin{tabular}[c]{@{}c@{}}action \\ specific\end{tabular} \\ \midrule
UADFV~\cite{yang2019exposing}                                                            & 49       & 49       & face swap                                                                          & No                                                         \\ \midrule
Celeb-DF~\cite{li2020celeb}                                                           & 590      & 5,639    & face swap                                                                          & No                                                         \\ \midrule
DFDC~\cite{dolhansky2020deepfake}                                                               & 48,190   & 104,500  & face swap                                                                          & No                                                         \\ \midrule
FF++~\cite{rossler2019faceforensics++}                                                              & 1000     & 4000     & animation \& swap    & No                                                         \\ \midrule
ForgeryNet~\cite{he2021forgerynet} & 91,630   & 121,617   &  animation \& swap                                                                    & No     

\\ \midrule
\textbf{\begin{tabular}[c]{@{}c@{}}DeepFake\\ MNIST+\end{tabular}} & 10,000   & 10,000   & image animation                                                                    & Yes     
\\ \bottomrule
\end{tabular}
}

\caption{Basic information of existing DeepFake datasets.}
\label{table:dataset_info}
\end{table}

\begin{figure}[]
\centering
\includegraphics[width=0.43\textwidth]{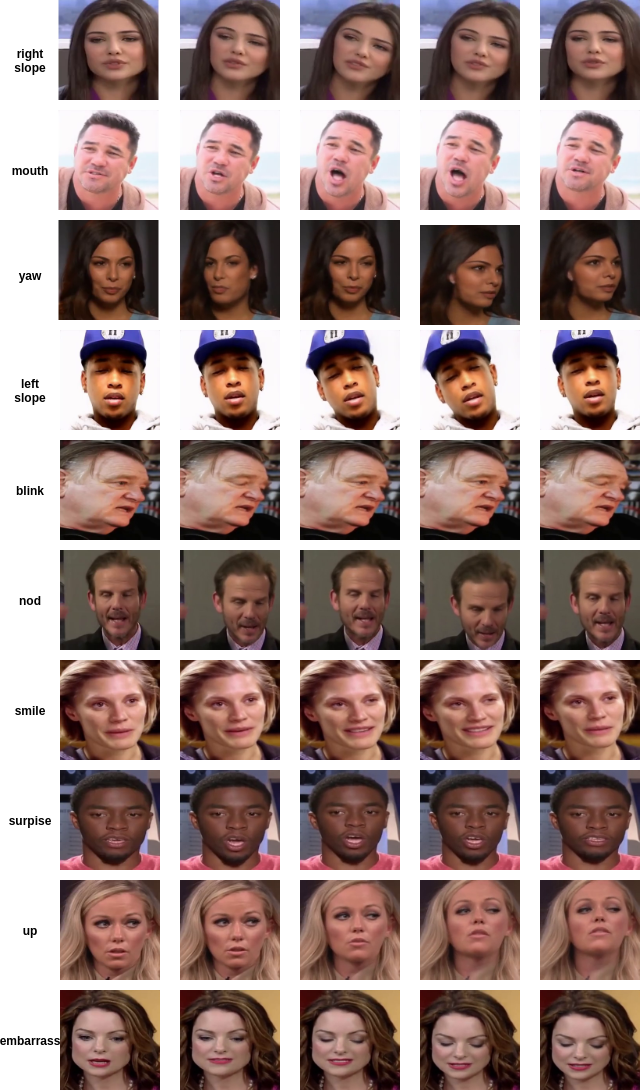}
   \caption{Facial animation video samples for different actions.}
\label{fig:short}
\vspace{-20pt}
\end{figure}

\vspace{-10pt}
\section{DeepFake MNIST+}

The major contribution of this paper is our proposed human face animation video dataset, called DeepFake MNIST+. It includes 10,000 face animation videos performing ten different actions and 10,000 real human face videos selected from other datasets. Besides, all these animation videos can spoof the liveness detection solution in the market. Such that the videos are still challenging recent public detectors. It is the first large-scale dataset for face animation videos of variant actions to the best of our knowledge. We believe such a dataset allows us to train advanced detection models to distinguish the face animation videos for preventing spoofing.

% \begin{figure*}[h]
% \centering
% \includegraphics[width=\textwidth]{LaTeX/images/animation_samples.png}
%   \caption{Forgery video samples for head left slope and open mouth actions.}
% \label{fig:short}
% \end{figure*}

\subsection{Generation Model and Data Preparation}
We select Siarohin's framework ~\cite{siarohin2020first} to generate face animation videos. This SOTA animation framework taking as input a source identity face image and a driving video shot by another actor. The model performs local affine transformations using first-order Taylor expansion to estimate the motion of the driving video. Then applying the motion features to the generator to provide high-quality face animation videos. As a result, it generates a video that animates the motion of driving video while keeping the identities of the source image. For more detail, please review the original paper. The major advantage of this framework is that the model is not identity-dependent. We can generate different videos for variant identities with arbitrary driving in the one trained model, while it only requires one image as the source image.

For the source identity images, we select the frames from video in VoxCeleb1 dataset~\cite{nagrani2020VoxCeleb}. VoxCeleb1 is a large-scale audio-visual dataset of human speech. It includes 1251 unique celebrities in 22,496 talking videos. All videos are face-cropped and have size 256x256 resolution. During our experiments, using the front face source images could achieve better generation quality. Therefore, we select the face frames mostly facing the front from the VoxCeleb1 dataset as our source images for generation.

% \subsection{Action Driving Videos}
The DeepFake MNIST+ dataset contains forgery videos in 10 actions. It includes: \emph{Blink, Open mouth, Yaw, Nod, Right slope head, Left slope head, Look up and smile, surprise and embarrassment}.
\vspace{-2pt}
The driving videos of embarrassment are collected from ADFES dataset~\cite{hawk2008moving}. The dataset contains variant emotional expression videos (anger, disgust, fear, joy, sadness, surprise, contempt, pride and are shot by 22 actors. We select five actors' embarrass videos from the dataset as our driving video. The videos of the remaining actions are shot by one volunteer. The action videos are captured using a front camera of the iPhone 11 Pro, and each action has been executed by the volunteer for 5 times. All the driving videos are face cropped by using MTCNN modules~\cite{zhang2016joint} and resized into 256x256 resolution to align the format of the VoxCeleb1 dataset.

\subsection{Generating High-Quality Facial Animations}
We adopt two public liveness detection APIs to select the challenging samples that cannot be accurately recognized by the detector. The first one was provided by TianyanData~\cite{38:online}. TianyanData's API supports liveness detection with a specific action, including blinking, yawing, nodding, and opening mouth. In order to pass the detection, the input face video has to perform a particular action while passing the spoofing test. The second one comes from Baidu~\cite{62:online}. Their detector supports universal liveness detection regardless of actions. Both of these two companies claim their detector can achieve 99\% accuracy for detecting spoofing.

We generate many animation videos for all actions and then pass the data to the liveness detection APIs to pick out the samples that are challenging for the liveness detector. As a result, the DeepFake MNIST+ dataset contains 10,000 face animation videos in 10 specific actions and 10,000 real face videos collected from VoxCeleb1. Each action includes 1,000 videos, and all of them can spoof the APIs. For blinking, yawing, nodding, and opening mouth, we use both TianyanData and Baidu APIs to filter videos. For the remaining actions, we use Baidu's API to collect spoofed data. The following graph presents each action's spoof rate by passing the animation videos to the two APIs.

% \begin{table}[H]
% \caption{The performance of XceptionNet models trained with existed datasets from FF DF+~\cite{rossler2019faceforensics++}. All models approach 100\% accuracy in their own datasets.}
% \centering
% \scalebox{0.75}{
% \begin{tabular}{c|ccc}
% \toprule
%                 & deepfake~\cite{GitHubdf58:online} & NeuralTextures~\cite{thies2019deferred} & Face2Face~\cite{thies2016face2face} \\ \midrule
% DeepFake Mnist+ & 43.7\%   & 63.6\%         & 67.5\%   \\
% \bottomrule
% \end{tabular}
% }
% \label{table::DeepFake_detector}
% \end{table}
% \vspace{-10pt}

The actions of blinking, yawing, nodding and opening mouth have lower average spoofing rates than others, such the situation could be caused by two APIs filter the videos of those actions. In addition, the TianyanData's API requires further action detection, which reduces the chance to attack. On the other hand, the actions that require large-angle head movement, e.g., yawing and nodding, the success spoofing rates are much lower than the other actions that don't need a significant motion change. One reasonable explanation could be that a single source image of the frontal face cannot provide sufficient detail of all head information, e.g., profile face, leading to lower head movement quality. The videos of simile have the highest spoofing rate, which has achieved 61\%. It might because the smile action doesn't lead to significant head change, making it hard to detect the spoofing details. While the yaw videos are more likely to be detected, that has a 23\% successful rate only.

\begin{table}[ht]

\scalebox{0.68}{
\begin{tabular}{c|cccc}
\toprule
                                                                   & DeepFake Mnist+ & deepfake & NeuralTextures & Face2Face \\\midrule
original accuracy                                                    & 96.58\%         & 99.7\%   & 99.2\%         & 98.9\%    \\ \midrule
accuracy on DM+ &       -          & 43.7\%   & 63.6\%         & 67.5\%    \\\midrule \midrule
fine-tuning                                                        & 95.3\%          & 98.46\%  & 98.1\%        & 98.49\%  \\\bottomrule 
\end{tabular}
}
\caption{The performance of Resnet50 models trained with existed datasets from FF++~\cite{rossler2019faceforensics++} to detect our proposed dataset. And the performance of DeepFake Mnist+ trained model, fine-tuning with FF++ data, to detect all these datsets.}
\label{table::existed_dataset_acc}
\end{table}

In addition, we explore the transferability of the detector trained with existing datasets. We train Resnet50~\cite{he2016deep} models with the data of deepfake~\cite{GitHubdf58:online}, Face2Face~\cite{thies2016face2face} and NeuralTextures~\cite{thies2019deferred} provided by FF++~\cite{rossler2019faceforensics++} and present the result for detecting DeepFake Mnist+. The first one is the face-swapping data. The last two are the expression manipulation dataset. The Table~\ref{table::existed_dataset_acc} presents the result. All these three models are trained with raw quality and achieve nearly 100\% accuracy in their own dataset. The result shows that the face-swapping detector fails to distinguish our proposed dataset. It seems better in manipulation datasets but still has a huge gap compared with the performance in their own datasets. Furthermore, we fine-tune the DeepFake Mnist+ trained model with the three FF++ datasets. The result indicates that the model can gain the power to detect both animation data and other Deepfake data by using our proposed dataset with other data sets. 

Based on the these results, we believe the current detectors still cannot defend against such attacks. These observations engage us to proposed a face animation dataset to improve the detectors and achieve better security. 
\begin{figure}[th]
% \centering
\includegraphics[width=0.47\textwidth]{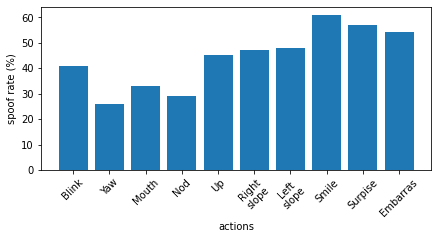}

\caption{The spoofing successful rates for different actions. The first four actions' videos are detected by TianyanData~\cite{38:online} and Baidu~\cite{62:online} API, while the later six actions' videos are detected by Baidu API.}
\label{fig::pass_rate_chart}\vskip -0.2in
\end{figure}

\section{Benchmark Systems}

We proposed a simple detection pipeline to detect the forgery videos in our dataset and distinguish them from those real videos. The forgery detection can be formulated as a binary classification task. We take the 10,000 real videos in the dataset as positive, while those forgery videos as negative.

% During the training process, we extract 16 frames for each video as the training samples.

In practice, the videos are usually suffering compression before uploading to the Internet, or the video might be shot by poor camera equipment, which suffering low video quality. Different compression rates are considered in our experiment to simulate the realistic detecting setting under different video qualities.  We select two different compression rates - C23 and C40 under H.264 codec to compress the videos. To be mentioned that, a higher compression rate indicates worse video quality.

We exploited multiple models to accomplish the DeepFake detection task in our experiment:

 \textbf{MesoInception-4}: MesoInception~\cite{afchar2018mesonet} is a CNN model consisting of two inception modules inspired by InceptionNet ~\cite{afchar2018mesonet}. The model uses mean squared error between true and predicted labels rather than the ordinary cross-entropy loss. Following the training procedure in ~\cite{rossler2019faceforensics++}, we extracted the frames as the original size which is 256x256 resolution. 

\textbf{XceptionNet}: The XceptionNet~\cite{chollet2017xception} is a traditional CNN model based on separable convolutions with residual connections. The model has shown high accuracy when detecting deepfake~\cite{GitHubdf58:online} videos and has been the baseline model introduced by Rossler et al.~\cite{rossler2019faceforensics++}. We used a pre-trained model on the ImageNet~\cite{deng2009imagenet} dataset in this experiment. The CNN layers are frozen, and we only update the weights of newly inserted full connected layers for the prediction. Same with MesoInception, we also used the 256x256 resolution of frames as the input.

\textbf{Resnet}: The Residual Neural network (Resnet)~\cite{he2016deep} is one of the most popular neural networks. It utilizes skip connections or shortcuts to jump over some network layers, such that the networks are easier to be optimized even with the increasing depth. The Resnet networks are also pre-trained with ImageNet~\cite{deng2009imagenet}. Three versions of Resnet - Resnet50, Resnet101, and Resent152 are included in the experiment to explore the performance change under different network depths. For all versions of Resnet, we capture and resize the image frames into 224x244 resolution.

All CNN models are trained with Adam optimizer with an initial learning rate of \(0.0002\), and we set \(\beta_1\) = 0.999 and \(\beta_2\) = 0.9. The learning rate decreases under the poly-decay schedule. The total number of training epoches for each model is set as 50, and the batch size is 64. We pick 70\% of videos from the DeepFake MNIST+  as the training data, that is, 700 videos from each of the ten action categories and 7000 real videos. 15\% videos are the validation data. The remaining 15\% will be the test data. We select the best versions of models based on the accuracy on the validation set. 

% \begin{figure}[h]
% \centering
% \includegraphics[width=0.47\textwidth]{LaTeX/images/dataset_detection_pipeline.png}
% \caption{The baseline detection pipeline for our proposed dataset.}
% \label{fig::dataset_detection_pipeline}
% \vspace{-10pt}
% \end{figure}

\vspace{-6pt}
\section{Evaluation}
This section presents different models' performance changes for detecting our proposed face animation video dataset under different situations. We present different models' accuracy for classifying test video data frames as real or fake ones to show the performance.
\vspace{-5pt}
\subsection{Overall Detection Performance}
\begin{table*}[h!]
\centering
\caption{The accuracy of different classifier models in testing set under different compression rate. The \textbf{light compression} corresponding to the c23 compression rate, while the \textbf{heavy compression} corresponding to the c40 compression rate. We select the best version of models based on the validation accuracy during the training process.}
\begin{tabular}{c|ccccc}
\toprule
                  & Resnet50 & Resnet101 & Resnet152 & XceptionNet & MesoNet \\ \midrule
raw               & 96.58\%     & 96.64\%   & 96.18\%   & 92.38\%     & 60.39\%    \\ 
light compression & 94.32\%  & 94.87\%   & 94.90\%   & 85.52\%     & 58.58\%        \\ 
heavy compression & 91.49\%  & 90.44\%   & 91.27\%   & 83.143\%    & 57.90\%         \\\bottomrule
\end{tabular}

\label{table::performance_results}
\end{table*}

\begin{table*}[h!]
\centering
\begin{tabular}{c|cccccc}
\toprule
          & Raw -\textgreater LC & Raw -\textgreater HC & LC -\textgreater Raw & LC -\textgreater HC & HC -\textgreater Raw & HC -\textgreater LC \\ \midrule
Resnet50  & 85.52\%               & 71.3\%                & 95.16\%               & 82.98\%            & 59\%                  & 61.02\%               \\
Resnet152 & 79.23\%               & 68.12\%               & 82.33\%               & 73.60\%               & 76.72\%               & 76.69\%               \\ 
Xception  & 79.17\%               & 71.83\%               & 87.89\%               & 68.99\%               & 67.38\%               & 62.9\%                \\ 
MesoNet  & 51.37\%               & 56.92\%               & 57.01\%               & 49.53\%               & 55.65\%               & 58.64\%                \\

\bottomrule
\end{tabular}
\caption{The models' performances trained by one specific compression rate and detecting the videos from the datasets of the other two compression rates. }
\label{table::performance_results_other_rates}
\vspace{-15pt}
\end{table*}

Table~\ref{table::performance_results} compared the accuracy of different models under three different video compression levels (raw, light and heavy compression). The result indicates that the Resnet models have the best performance among all models. They achieve a 96.3\% average accuracy on the raw video dataset, which is much higher than those of the other two CNN models. The XceptionNet, that also has a deep architecture, approaches a 92.38\% accuracy for detecting forgery videos. In addition, the MesoNet shows the poorest performance with only a 60.39\% accuracy for detecting the raw videos. The increasing of the network depth does not constantly boost the performance. The accuracy of three Resnet variants are very similar. It is hard to say, the deeper architectures, e.g., Resnet152, show improvement compared to other smaller models.

The video quality is also an important factor affecting the performance. Table~\ref{table::performance_results} shows that worse video quality (higher compression rate) could lead to a performance downgrade. Compared to the raw dataset, the average accuracy of Resnet networks decreases around 2\% under the light compression condition. The situation is worse when the videos are under heavy compression. 91.06\% of compressed videos are classified correctly with Resnet networks on average, which has a 5\% gap from that on the raw dataset. The XceptionNet is more sensitive to compression rate than other networks, whose accuracy drops to 85.52\% significantly when applying light compression and 83.143\% for heavy compression.

The decreasing correction rate could be caused by the loss of detail in low video quality. The compression process leads to blur frames, hiding the forgery information so that the detector might not be able to capture such information for detecting the forgery areas.

\subsection{Analysing the Impact of Video Quality}

We further explore how the video quality could affect the performance. The Table~\ref{table::performance_results_other_rates} shows the models' generalization for the videos of different compression rates. It presents the accuracy of models trained with one quality (e.g., Raw) and predicts the data with the other two qualities(e.g., light and heavy compressed). The results indicate that the models cannot adapt well to datasets with other qualities, especially between raw and heavy compression datasets. The raw video models only achieve 70\% accuracies on average for heavy compression videos and 67\% conversely. It might show the heavy compression could change the data distribution leading to a dramatic accuracy drop. Also, light compression models have better generalizations than others, indicating the models could learn animation and compression information simultaneously.

One way to overcome the impact of different video quality is to train the network with the videos in all qualities. We train the Resnet50 and Resnet152 with the mixed video quality dataset for the experiment. More specifically,  we select the video quality randomly for each sample to train the models and present their performances in testing sets under three qualities respectively. The Table~\ref{table::data_mixed} shows the result. With training in mixed quality videos, the models can adapt to different video qualities. However, it still has a slight accuracy decrease, especially for heavy compression videos. This change supports our previous observation, which could have a large difference in distribution between raw and heavy compressed videos. The result also suggests that it might require a large model to learn the mixed video quality dataset. Resnet152, a deeper network, has smaller gaps with the models trained with a single-quality video set, which has 3\% improvement for all qualities compare to the smaller Resnet50 network.  

% \vspace{-10pt}
\begin{table}[]
\centering
\begin{tabular}{c|cccc}
\toprule
                  & Resnet50 & Resnet152 & XceptionNet & MesoNet\\ \midrule
Raw               & 93.57\%  & 95.78\%   & 90.82\% & 59.47\%\\ 
LC & 90.69\%  & 92.11\%   & 83.28\% & 56.94\%\\ 
HC & 85.56\%  & 88.32\%   & 82.43\% & 55.38\%\\ \bottomrule
\end{tabular}
\caption{The performances of models trained with mixed video quality dataset in different testing sets.}
\label{table::data_mixed}
\vspace{-10pt}
\end{table}

\subsection{Evaluation of the Training Corpus Size}

We evaluate how the training corpus size could affect the detection performance. We select 10000, 3000, 1000, 500, 100, and 10 animation videos and the corresponding number of real videos to train the Resnet50 model. The animation videos for each action are selected equally. 

\begin{figure}[h]
\centering
\includegraphics[width=0.47\textwidth]{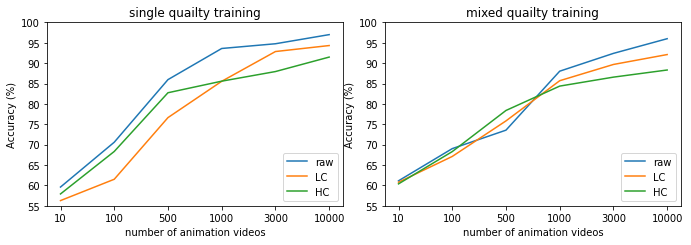}
\caption{The performance changes under different training corpus sizes using single quality or mixed quality video dataset.}
\label{fig::performance_size}

\end{figure}

The chart on the left of Figure~\ref{fig::performance_size} presents the importance of training corpus size. It could only lead to a small downgrade in the raw dataset when keeping 10\% of the data, but more impact when the videos are compressed. It could indicate that more low-quality videos are required to train the models.  Besides, correction decreases dramatically if we use 1\% of data for training in all quality. Simultaneously, the models tend to random guessing when we only have ten animation videos for training.

\begin{figure}[h]
\centering
\includegraphics[width=0.42\textwidth]{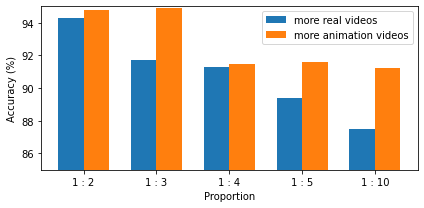}
\vspace{-10pt}
\caption{The detection accuracy of Resnet50 models trained with raw videos under different proportions of real and animation videos.}
\label{fig::video_propotion}
\vspace{-10pt}
\end{figure}

In addition, Figure~\ref{fig::video_propotion} shows performances under different proportions of animation and real data for training.  More specifically, we reduce the number of one class's videos (either real or animation) to reach the expected proportions of videos for training. For example, "1:2 more real videos" means we remove half of the animation videos for training. Similar to decrease the total training corpus size, either reducing the real or animation videos will lead to an accuracy decrease. Also, preserving animation videos for training is more critical than introducing more real videos. When we use 10\% of animation video for training, the performance only achieves 85\%; a 10\% decrease compares to the full dataset. On the contrary, keeping 10\% of real videos with full animation videos leads to a smaller 5\% downgrade.  
    
The chart on the right of Figure~\ref{fig::performance_size} presents how the training corpus size could affect the performance of models trained with the mixed quality dataset. We trained the Resnet152 models with 3000, 1000, 500, 100, and 10 animation videos and the corresponding number of real videos under mixed video quality. Similar to the models trained with single quality datasets, a smaller corpus size also reduces corrections for the mixed quality situation. We also notice that the performance will slightly lower for large training corpus size than single-quality videos trained models. However, when the corpus size becomes smaller, the accuracy tends to be higher than the model trained with single-quality videos, especially for light compression videos, which suffering the most impact relate to decreased corpus size. One reason could be that the mixed video quality training strategy is similar to data augmentation, which increases the data diversity for small training corpus size to improve performance.

\vspace{-10pt}

\subsection{Evaluating the Impact of Type of Actions}

\begin{figure}[h]
\centering
\includegraphics[width=0.42\textwidth]{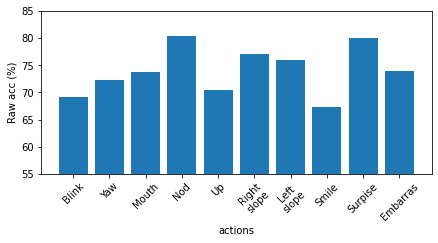}
\vspace{-12pt}
\caption{The detection accuracy for the Resnet50 models trained with one specific actions videos. Each label indicates the model of the whole testing data performance, trained with the videos of that specific action and 1,000 real videos.}
\label{fig::each_action_train_only}
\vspace{-10pt}
\end{figure}

The type of actions could affect the detection performance. We train the models with videos of one single action and evaluate whether the models could adapt to other unseen animation videos. For each action, we select all animation videos of that action and the corresponding number of real videos (1000 videos for each label) to train the Resnet50 models and test the performance with the whole animation testing video set. We select the raw quality videos for the experiment, and Figure~\ref{fig::each_action_train_only} shows the result. With the training of single-action, it is not doubted that models cannot keep similar performance to the one trained with full actions raw quality videos with 1000 corpus size. The accuracy drops to 74.3\% on average from 93.4\%. The videos of nodding and surprising actions provide a better generalization to adapt unseen actions, which achieve 80.46\% and 79.98\% respectively. Right and left slope videos also introduce relatively high performance than remaining actions, which reach 77.02\% and 75.89\% respectively. On the other hand, the models trained with smile and blink videos have poor correction rates, which drop to 67.36\% and 69.08\%. In summary, using large movement action videos to train classifiers could lead to better performance for detecting new actions' videos. The actions that only include small changes, e.g., smiling and blinking, cannot provide sufficient information for the networks to adapt to unseen videos.

\begin{figure}[h]
\centering
% \vspace{-10pt}
\includegraphics[width=0.42\textwidth]{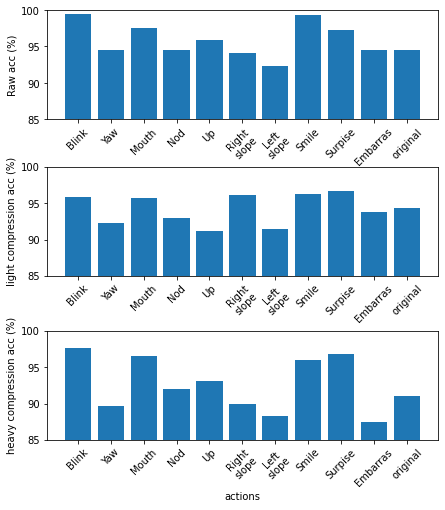}
\vspace{-12pt}
\caption{The detection accuracy for each action under different video quality. The models are trained with full training dataset. The original class indicate the selected real videos.}
\label{fig::each_action_merge}
\vspace{-15pt}
\end{figure}

We also compare the full dataset trained models' performances for detecting different actions under three video qualities, and Figure~\ref{fig::each_action_merge} shows the result. We can notice that some actions' videos are relatively hard to detect in all video qualities. The left slope videos are the most difficult ones, which the accuracy is 92.3\% under raw videos and drop to 88.24\% under heavy compression videos. In addition, the detection of embarrassment videos might require higher video quality. Its performance decreases to 87.5\%, the lowest one under heavy compression and a huge 7\% gap compare to the one in raw quality. On the other hand, some actions, e.g., smiling, blinking, are much easier to be detected with the networks. They have achieved 100\% correctness for raw videos, decrease to 97\% on average if the videos are heavily compressed, which still keep in a higher level than other actions. 

Comparing with Figure~\ref{fig::each_action_train_only}, we observe that some hard-to-detect actions, e.g., right and left slope, could provide more generalization if we only use those actions' videos for training. On the contrary, the models trained with the videos of easy-to-detect actions, e.g., smiling and blinking, showing poorer performances for adapting unseen actions.

% \vspace{-5pt}

\subsection{Visualizing the Attention Parts of Models}
We analyze what the classifier learned for distinguishing the animation and real videos. More specifically, we try to visualize the network pay attention to which part of the video frames for detection. We use Class Activation Map (CAM)~\cite{zhou2015cnnlocalization} to achieve visualization. The CAM is a technique to visualize what the classifier is looking at. CAM relies on the output from the models' global average pooling(GAP) layer, which right after the last convolution layer. The GAP layer could keep the spatial information of the convolution layer. By multiplying the weight of the Softmax layer of one specific class with GAP output, we can visualize the models' attention regions for classifying the given input images as that specific class. Our baseline models are binary classifiers to separate frames from either real and animation videos. In this case, the CAM results present semantic information about which parts of input image are critical regions for the models to decide whether it is from the animation videos.

\begin{figure}[h]
\vspace{-10pt}
\centering
\includegraphics[width=0.35\textwidth]{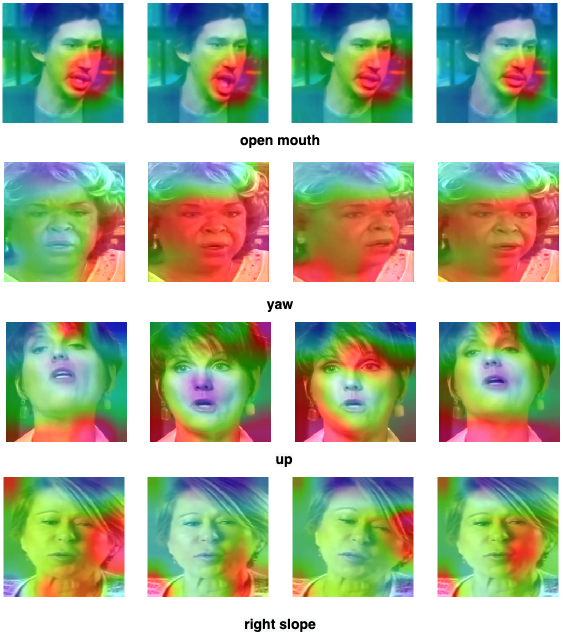}
   \caption{The Class Activation Map (CAM) results of the Resnet50 model trained with raw videos.}
\label{fig::action_maps}
\vspace{-15pt}
\end{figure}

In our experiment, we visualize the CAM results of the Resnet50 model trained with raw quality videos. To be noticed that, the GAP layer has been added to the model in the original design, so we don't require to do the further modification. The Figure~\ref{fig::action_maps} demonstrates some CAM results of selected video frames of different actions. We adapt the results to the original images to highlight the attention parts. The red color region indicates the important region, while the model pays less attention to the blue color areas. The results indicate that the model could learn semantic information to detect the animation videos. The model relies on the forgery regions to make the decision. For the opening mouth video frames, the model focuses on the detailed information of mouth. Similarly, the head and neck regions of the frames could be significant for detecting up videos. And the network pays attention to the profile face for head movement action's video frame, like yawing and sloping.

\section{Discussion and Conclusion}
We present a new large-scale, action-specified facial animation video dataset - Deepfake Mnist+, and evaluate the dataset's detecting performance with the proposed baseline detection method in different situations. We mainly explore the impact of the compression and actions on classification accuracy. It indicates the low-quality videos could significantly affect the performance and large movement actions could provide further generalization for unseen data. We expect that our proposed dataset could improve the detection performance of facial animation videos and increase the robustness and security of the recent liveness detectors as a response to the concern about pervasive animation videos online. As future work, we will explore other advanced facial animation methods and enlarge our datasets with more actions shot by more actors.

\clearpage
{\small
\bibliographystyle{ieee_fullname}
\bibliography{egpaper_final}
}

\clearpage
\appendix
\section{Data Generation And Collection Pipeline}
\begin{figure*}[b]
\centering
\includegraphics[width=\textwidth]{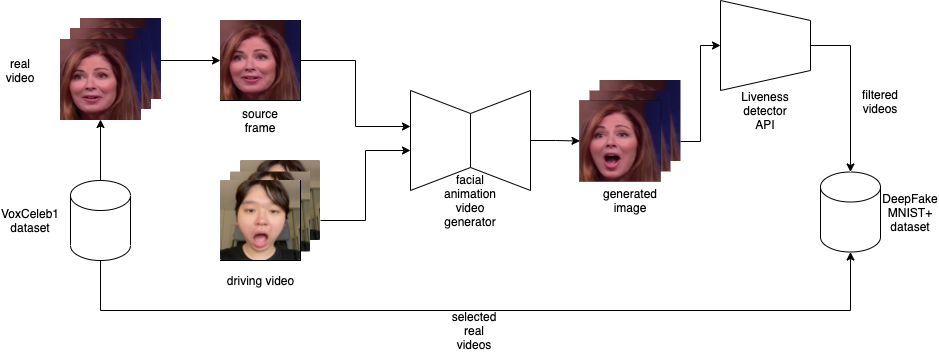}
  \caption{The pipeline to generate and collect our proposed DeepFake MNIST+ dataset.}
\label{fig::pipeline}
\end{figure*}

% \clearpage

The Figure~\ref{fig::pipeline} shows the pipeline to generate and collect our proposed DeepFake MNIST+ dataset. First, we collect the real videos from the VoxCeleb1~\cite{nagrani2020VoxCeleb}, extract the frames from these real videos as the source identity images. Then we shot driving videos with ten actions through the volunteers. In order to match the format of VoxCeleb1 videos, which are face-cropped and have the size of 256x256 resolution. We made a face-cropped version of driving videos with MTCNN modules~\cite{zhang2016joint} and also resize them into 256x256 resolution. We use Siarohin's framework~\cite{siarohin2020first} for animation video generation to produce face animation videos. It is a SOTA animation framework such that it even could capture the detail of eyeball moving. A single source image and a driving video were passed to the generator each time to produce single animation videos with a specific action. The generated videos were filtered with Liveness detector APIs to collect the challenging videos. Finally, 10,000 passed animation videos with ten actions (1000 videos for each action) and selected 10,000 real videos from VoxCeleb1 from our proposed DeepFake MNIST+ dataset.

\section{More Examples of Data}
\subsection{Compressed Images}
\begin{figure*}[!ht]
\centering

\includegraphics[width=\textwidth]{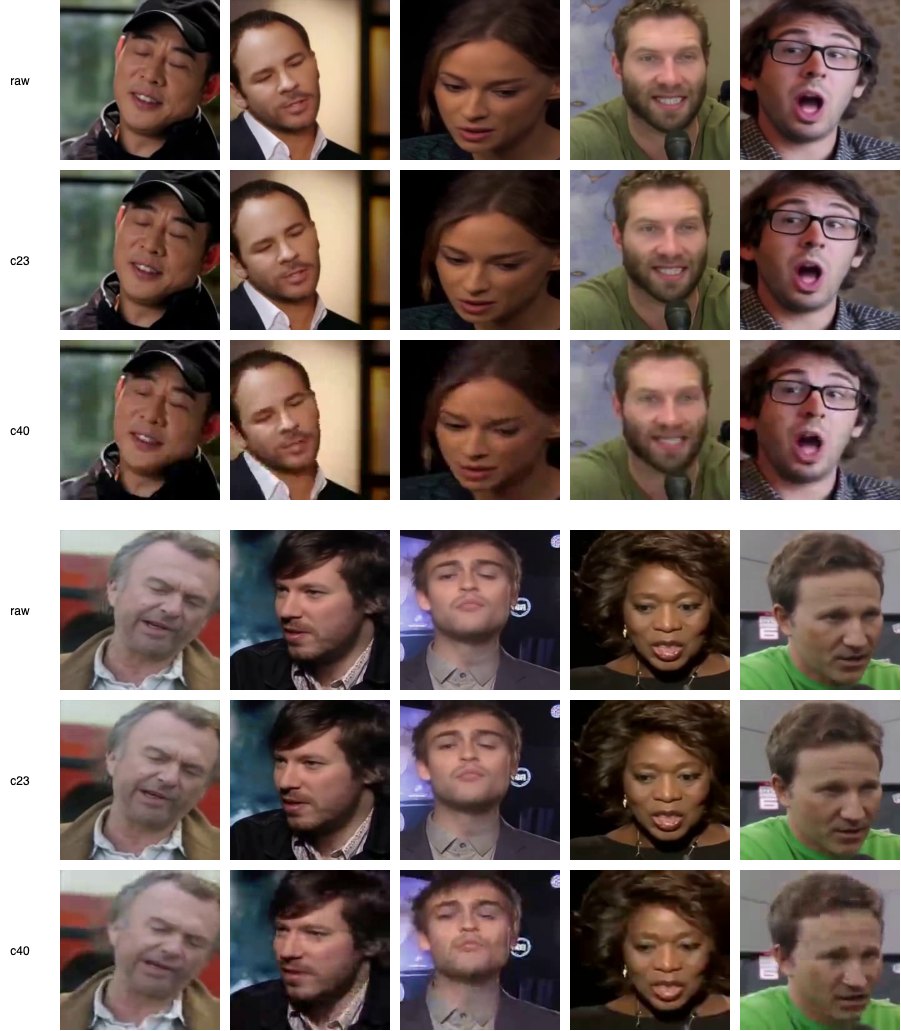}
  \caption{The samples of raw and corresponding compressed (both c23 and c40) video frames. }
  
\label{fig::compressed}

\end{figure*}

We demonstrate some raw and compressed (in both C23 and C40 compression rate under H.264 codec) video frames in Figure~\ref{fig::compressed}. The higher rate means heavier compression. As we can see, the c23 compression rate only leads to a minor impact on visual video quality. However, the frames suffer significant detail loss and blur effect under the c40 compression rate.

\subsection{Driving Video samples}
\begin{figure*}[!ht]
\centering

\includegraphics[width=0.7\textwidth]{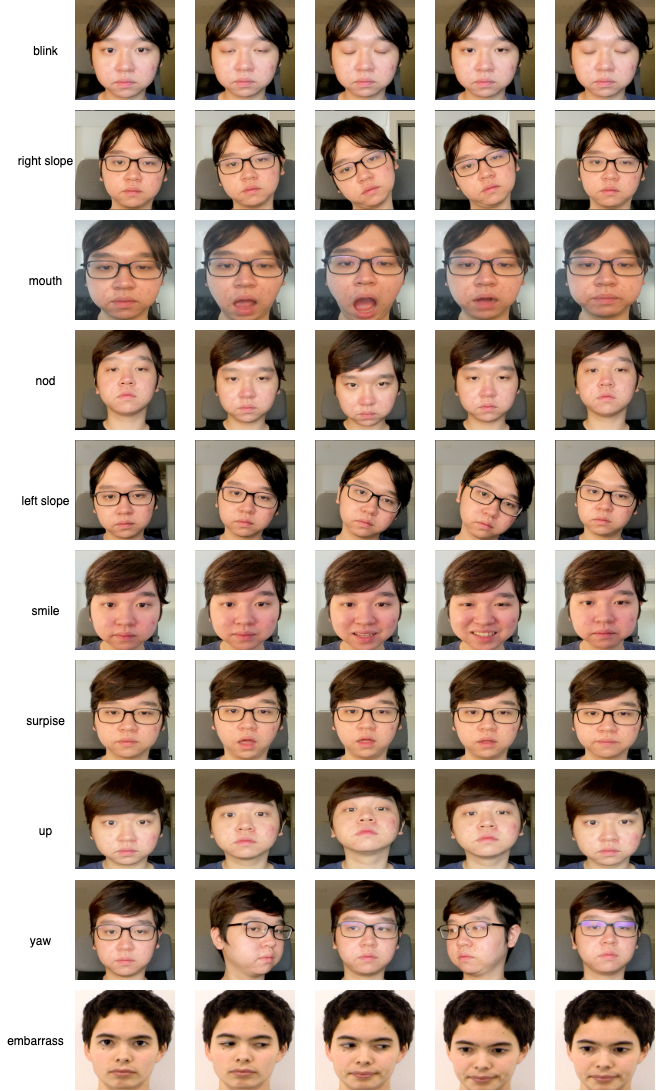}
  \caption{The driving video samples for animating the source images.}
  
\label{fig::driving}

\end{figure*}

In this section, we present some driving video samples of different actions for animating the source images in Figure~\ref{fig::driving}. The driving videos of embarrassment are picked from ADFES dataset~\cite{hawk2008moving}. 

\clearpage
\section{More experiments}
\vspace{-10pt}
\begin{table}[h]
\centering
\begin{tabular}{|c|c|c|c|}
\hline
                 & DFDC\cite{dolhansky2020deepfake}   & DF-1.0\cite{jiang2020deeperforensics} & Celeb-DF\cite{li2020celeb} \\ \hline
without finetune & 61.2\% & 60.7\% & 57.2\%   \\ \hline
finetune         & 95.6\% & 96.1\% & 95.1\%   \\ \hline
\end{tabular}
\caption{Accuracy of detecting DeepFake Mnist+ using the models trained with previous datasets. Fine-tuning means fine-tuned with proposed dataset.}
\label{table::performance}
\end{table}

% \newpage

\end{document}